\documentclass[journal]{IEEEtran}

\ifCLASSINFOpdf
\else
   \usepackage[dvips]{graphicx}
\fi
\usepackage{url}

\usepackage[nocompress]{cite}
\usepackage{cite}
\usepackage{hyperref}
\usepackage{graphicx}
\usepackage{amsmath}
\usepackage{amssymb}

\usepackage{CJK}
\usepackage{subfigure}

\usepackage[misc]{ifsym} 
\usepackage{bbding}
\usepackage{authblk}
\usepackage{float} 

\def\etal{\emph{et al}}

\begin{document}

\title{Self-Supervised Light Field Depth Estimation Using Epipolar Plane Images}

\author{Kunyuan Li ,  Jun Zhang\thanks{Corresponding author:zhangjun@hfut.edu.cn} ,  Jun Gao ,  Meibin Qi\\
Hefei University of Technology\\
Hefei, Anhui, China\\

{\tt\small lkyhfut@gmail.com, zhangjun@hfut.edu.cn, gaojun@hfut.edu.cn, qimeibin@hfut.edu.cn}

}

\markboth{Journal of \LaTeX\ Class Files, Vol. XX, No. X, 2021}
{Shell \MakeLowercase{\textit{et al.}}: Bare Demo of IEEEtran.cls for IEEE Journals}
\maketitle

\begin{abstract}
Exploiting light field data makes it possible to obtain dense and accurate depth map. However, synthetic scenes with limited disparity range cannot contain the diversity of real scenes. By training in synthetic data, current learning-based methods do not perform well in real scenes. In this paper, we propose a self-supervised learning framework for light field depth estimation. Different from the existing end-to-end training methods using disparity label per pixel, our approach implements network training by estimating EPI disparity shift after refocusing, which extends the disparity range of epipolar lines. To reduce the sensitivity of EPI to noise, we propose a new input mode called EPI-Stack, which stacks EPIs in the view dimension. This method is less sensitive to noise scenes than traditional input mode and improves the efficiency of estimation. Compared with other state-of-the-art methods, the proposed method can also obtain higher quality results in real-world scenarios, especially in the complex occlusion and depth discontinuity.
\end{abstract}

\begin{IEEEkeywords}
Light Field, EPI, Depth Estimation.
\end{IEEEkeywords}

\IEEEpeerreviewmaketitle

\section{Introduction}

\IEEEPARstart{L}{ight} field (LF) cameras~\cite{Ng} enable dense sampling of the viewpoints, which can collect both 2D spatial and angular information of the observed scene. Sub-aperture images and Epipolar Plane Images (EPIs)~\cite{Levoy} are common visualization ways of captured 4D light field data. Compared to the traditional 2D images, the 4D light field data provides information about multi-view and epipolar geometry, which makes it possible to estimate a dense and accurate depth map. In recent years, with the rapid development of deep learning technology, the performance of 4D light field depth estimation has been greatly improved. Generally, sub-aperture images and EPIs serve as the input of the CNNs, and multi-view and epipolar features are extracted to compute scene depth. Many works~\cite{Feng2018,EPI-volume,EPI-shift,EPI-ORM} exploiting these data patterns have been proposed. However, with the increase of input cost, the network also shows higher complexity and more training parameters. Moreover, existing datasets~\cite{dataset2016,dataset2013} with per-pixel ground truth are difficult to support training for large networks. Synthetic scenes are usually limited to a small 
disparity range. Training networks on synthetic datasets does not perform well in real-world scenes~\cite{EPI-shift}.

\begin{figure}[t]
\begin{center}
\includegraphics[width=1.01\linewidth, height=0.3\linewidth, scale=0.7,trim=12 0 15 0,clip]{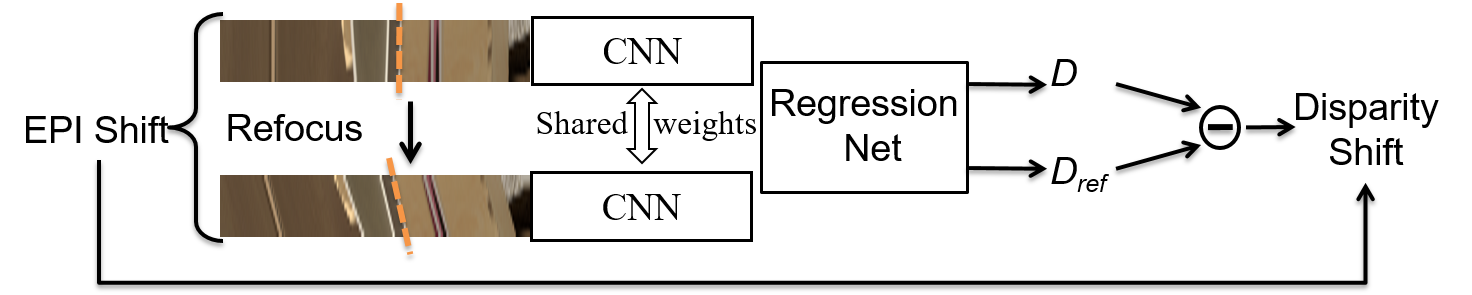}
\end{center}
   \caption{Depth estimation based on Self-supervised learning. This is illustrated with the EPI in the horizontal direction. The EPIs before and after refocusing are input to the network sharing weight. The difference between $D$ and $D_{ref}$ is the disparity shift determined by the EPI shift. Depth labels are not used here.}
\label{fig:self}
\end{figure}

Classical methods often use refocusing cues~\cite{Tao2013} for scenes to calculate scene depth. These methods can obtain high estimation accuracy even for real scenes with different disparity ranges. 
As a unique property of light field data, refocusing can change the depth plane of the scene, so that a single scene point contains different disparity maps, which shows the varying focus depth levels. 
Light field refocusing in depth estimation is mainly limited to traditional methods~\cite{Tao2017,Wanner2014,Williem2016} containing handcrafted parameters. It has less been explored that how to combine light field refocusing~\cite{focal-stack} and existing networks for depth estimation. 
Light field refocusing enables slicing of different depth planes in the same scene, which allows the limited number of datasets to provide rich focus cues. Consequently, in order to make full use of the existing datasets, it is necessary to fuse CNNs and refocusing principle to achieve more accurate depth estimation from light field, especially for real scenes. We prefer to implement network training without per-pixel ground truth. In this case, we expect that the network can still learn effective feature representation according to the changes of light field 2D slices before and after refocusing. Hence, inspired by~\cite{RankIQA,RankIQA-2019}, we consider the use of self-supervised learning to improve the training of network for depth estimation. The main idea (see Fig.~\ref{fig:self}) is that refocusing shifts can be directly estimated compared to the depth value calculated per pixel. By using appropriate auxiliary tasks, self-supervised learning can implicitly provide effective information for the original task. No annotation is required for auxiliary tasks. In this paper, we make full use of the existing dataset to improve the accuracy of depth estimation by computing the refocusing shifts. Considering the advantages of EPIs in complex texture and occluded regions, we use refocused EPIs as input. The light field refocusing extends the slope of epipolar line, which makes the network more suitable for various real scenes. 

In addition, to solve the problem of insufficient spatial information in perspective dimension of traditional 2D EPIs, this paper proposes to use EPI-Stack as input. Current EPI-based depth estimation methods usually only compute a single pixel~\cite{EPI-ORM,EPI-patch}, which is inefficient. This is mainly because when using multi-directional EPI for estimation, the common region of these input data contains only one pixel from the central view~\cite{Feng2018}. This also makes depth estimation using EPIs susceptible to noise in the scene, and it is difficult to extract effective linear structure features. EPI is the sampling of a certain space and perspective dimension. Due to the limitation of a single spatial dimension, it is easy to be affected by the local spatial position. 
Although using sub-aperture image as input can extend the common region~\cite{EPINET}, it is not effective to estimate some details and occlusions. In this paper, the input of EPI-Stack is used to increase the common region of EPIs in different directions. Our approach improves the performance of EPI in noisy scenes. 

The contributions of the paper are summarized as follows:

\begin{itemize}
\item[$\bullet$]
In this paper, a depth estimation framework based on self-supervised learning is presented. We build the corresponding auxiliary task, which is to estimate the refocusing shifts and improve the network training. This method has advantages over occlusions in complex real scenes.
\item[$\bullet$] 
We propose an input mode for EPI-Stack, which enables EPIs in different directions to have the same spatial resolution, higher efficiency of depth estimation, and better robustness to noise.
\item[$\bullet$] 
Compared with the existing state-of-the-art methods, the proposed method performs favorably on the synthetic and real scenes.
\end{itemize}

\begin{figure}[t]
\begin{center}
\includegraphics[width=1.05\linewidth, height=0.4\linewidth,trim=5 12 5 0,clip]{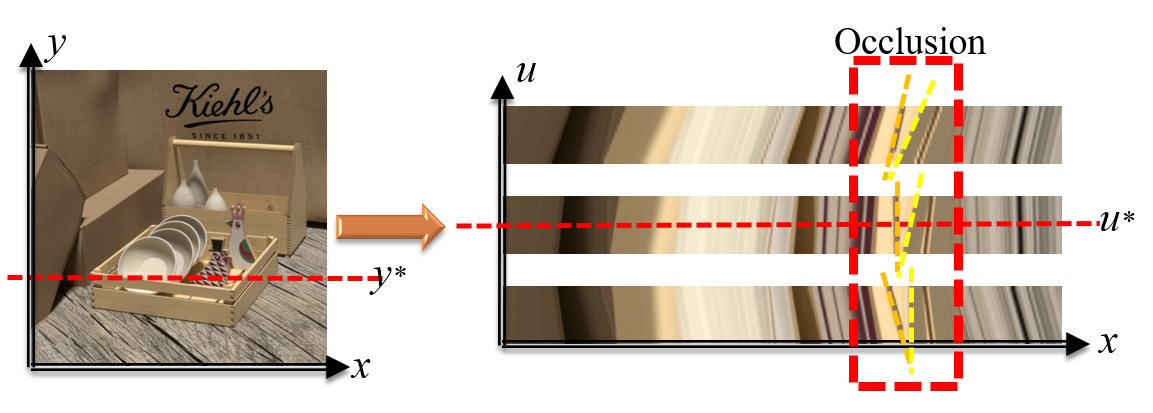}
\end{center}
   \caption{Refocused EPIs in occlusions.}
\label{fig:refocused}
\end{figure}

\section{Related Works}
In this paper, the related work is briefly introduced from two aspects of light field depth estimation and self-supervised learning.
\subsection{Depth Estimation from Light Field}
Depending on the types and depth clues of the input data, current methods of light field depth estimation can be divided into sub-aperture image-based, EPI-based and refocusing cues-based depth estimation. Traditional methods ~\cite{Jeon2019,Tao2017,Wang2015,Wanner2014,Williem2016}focus on the structural characteristics of the light field and obtain the depth map from the hand-crafted features. These methods are generally accompanied by complex calculation and subsequent optimization, and need to be manually tuned for different scenarios. Recently, by building an end-to-end network, deep learning technology has made a breakthrough in the application of pixel level light field, such as depth estimation, saliency detection~\cite{saliency}, super-resolution reconstruction~\cite{Super-Resolution} and so forth.

{\bf Methods based on Sub-aperture Images.} Inspired by the multi-view and stereo algorithms, Jeon \etal~\cite{Jeon2015,Jeon2019} use the sub-aperture images to estimate the multi-view stereo correspondences with a sub-pixel accuracy shift. These methods show good performance in the real scene captured by the lenslet-based camera, but suffer from noise at the occlusion boundary. Considering the narrow baseline between sub-aperture images, some methods~\cite{EPI-volume,EPI-shift,EPINET} use it as the input of CNN to estimate the disparity of the central view. On the basis of U-net~\cite{Unet2015}, Heber \etal~\cite{EPI-volume} propose the concept of EPI-Volume, using stacked sub-aperture images as input, which are located in the same horizontal or vertical perspective. Shin \etal~\cite{EPINET} extract four directional sub-aperture images in the epipolar plane and stacked them into the network to achieve the state-of-the-art performance. Using the input method in~\cite{EPINET}, Leistner \etal~\cite{EPI-shift} shift the stack of sub-apertures to make the network better adapt to the small- and wide- baseline light field. However, it is difficult to estimate the detail features in scenarios with this method.

{\bf Methods based on EPI.} Previous EPI-based depth estimation methods~\cite{Wanner2014,Wanner2013,SPO,EPI2017} prefer to design complex operation algorithms to calculate slope according to the structure of epipolar line. Although high estimation accuracy can be obtained, especially for occlusions, hand-crafted parameters and optimization are unavoidable, like most traditional methods. To overcome this, recent works~\cite{Intrinsics,Feng2018,Heber-shape,Heber-Unet,EPI-ORM,EPI-patch,Sun2016,EPI-Patch2018} have treated EPI as the input of CNNs for end-to-end estimation. Luo \etal~\cite{EPI-patch} designed a two-stream CNNs to estimate the depth of a single pixel from each patch. Zhou \etal~\cite{EPI-Patch2018} explore the estimation effect of different scale EPI patches and fused multi-directional features. In ~\cite{EPI-ORM}, Li \etal propose a relation network to extract orientation relation features between EPI center view and neighboring pixels, which can further improve the estimation accuracy. However, the current CNN-based method uses a single spatial dimension of EPI image and only one common pixel point in different directions, which makes it sensitive to noise.

{\bf Methods based on Refocusing cues.} Light field refocusing cues are widely used in traditional depth estimation methods. Most researches focus on exploiting different depth features, such as defocus, correspondence, shade and occlusion. Tao \etal~\cite{Tao2013,Tao2017,Tao2015} propose a depth estimation that combines defocus and correspondence features. Based on their works, Wang \etal~\cite{Wang2015} create occlusion-aware depth-maps via a modified angular photo-consistency. Lin \etal~\cite{focal-stack} synthesize the symmetry of focal stack and data consistency measure, which is more robust to noise and under-sampling. William \etal~\cite{Williem2016} also propose a data cost based on an angular entropy metric and adaptive defocus responses. These methods, which utilize the unique refocusing capability of the light field, can be adapted to complex and diverse real scenes, but also introduce a large number of hand-crafted features that are more complex. Current methods that combine refocusing with CNNs have rarely been studied, and the existing works usually only use the focused central view image as input, failing to learn the data changes before and after refocusing. For example, Zhou \etal~\cite{focal-cue} estimate the disparity label of each pixel from a set of focal stack with the continuous disparity range.

Overall, compared with traditional depth estimation methods, CNN-based ones can significantly improve the estimation accuracy. However, refocused 2D slices (sub-aperture images or EPIs) of light field data in different depth planes provide abundant refocusing cues, which cannot be effectively used by the existing CNN-based methods. This also makes it perform poorly in complex scenes, especially real scenes. In order to effectively use the light field focusing clue, in this paper we focus on building a unified training framework to fuse CNN with refocusing.

\begin{figure}[t]
\begin{center}
\includegraphics[width=1\linewidth, height=0.6\linewidth]{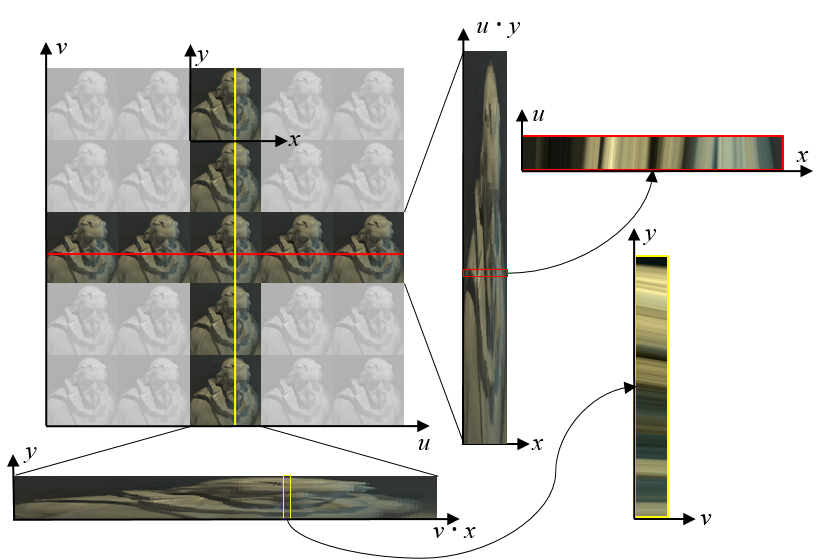}
\end{center}
   \caption{Illustration of the proposed EPI-Stack.}
\label{fig:EPI-Stack}
\end{figure}

\subsection{Self-Supervised Learning}
In recent years, self-supervised learning has attracted extensive attention because it provides an alternative method to label existing datasets. By constructing auxiliary tasks, self-supervised learning extracts corresponding supervised information from a large number of unlabeled data. This enables better training of deep neural networks. Liu \etal~\cite{RankIQA,RankIQA-2019} propose image ranking as an auxiliary task for some regression problems. Through the training of this task, the network can learn different semantic features. Then the network parameters are transferred to the original task for fine-tuning, which allows for deeper and wider training of the network. Other self-supervised tasks include predicting the relative position~\cite{Doersch} of patches in images, restoring the entire image from surrounding pixels~\cite{Pathak}, and generating color images from gray-scale images~\cite{Larsson}. 
In the field of monocular depth estimation, stereo matching clues are usually used to implement self-supervised training of network according to scene disparity and view position~\cite{Unsupervised-Depth2016,Unsupervised-Depth, mono-depth-dig,mono-depth}. This shows high estimation accuracy in both image and video sequences.

Inspired by these works, we propose a self-supervised depth estimation framework based on light field refocusing, which can be considered as the first work of light field depth estimation exploiting self-supervised learning. Different from the task of depth regression, this paper estimates the disparity shifts of scenes before and after focusing to get the refocusing cues, so as to learn the effective feature representation of 2D slices of light field data. Our approach not only improves network training, but also enables the network to better adapt to complex scenarios.

\section{Learning Disparity Shifts from Refocused EPIs}
\label{sec3}
Objects with different depths have different disparity corresponding to multi-view images. Based on this, light field refocusing can obtain images with different refocused depths. EPI is a visualization method for two-dimensional slices of light field data. The slope of the epipolar line contains the disparity information of the scene. Compared with sub-aperture images, EPI has more significant structural changes before and after refocusing. As shown in Figure \ref{fig:refocused}, for the same scene point, the slope of the epipolar line varies depending on the refocused depth. Due to the different depth of foreground and background, the polar slope in the occlusion also shows a significant difference. The unique linear structure of EPI can provide more effective refocusing cues, and it has advantages for complex occlusions to achieve more accurate depth estimation. Therefore, this paper designs the corresponding CNN to capture the change of epipolar slope before and after refocusing. Unlike the current method of directly using the refocused EPI as input, we fuse the light field refocusing with CNN in a self-supervised learning manner. As shown in Figure \ref{fig:self}, each pair of EPI images before and after refocusing is input into the same CNN separately, and the weights are shared for feature extraction. Within the disparity range of a scene, the same scene point can be refocused multiple times to obtain different EPI pairs, which expands the number of samples. Thus, this self-supervised learning method can learn more effective feature representations from a limited set of dataset and promote network convergence. Then we can get the disparity $D$ and $D_{ref}$, which are the disparity values before and after refocusing, respectively. According to the mapping relationship between the polar slope and disparity~\cite{Wanner2014}, the disparity value is shown in formula~\ref{eq:DEF}. 
Here, we use the $u_0$ central perspective as a reference and assume the same baseline $\Delta u$ between adjacent views. $Z$ and $f$ represent scene depth and focal length, respectively.
The difference between the output disparities should be consistent with the EPI shifts after refocusing. The EPI shift $E_{s}$ can be predefined as the Ground Truth of the network for end-to-end estimation. Therefore, the loss function between the $E_{s}$ and the difference of output disparity is shown in formula~\ref{eq:LOSS}. 

\begin{equation}
{D_{ref}} - D = ({u_0} - u)\frac{{\Delta u}}{Z}f
\label{eq:DEF}
\end{equation}

\begin{equation}
Loss = \frac{1}{m}\sum\limits_{i = 1}^m {\left| {(D_{ref}^i - {D^i}) - E_s^i} \right|} 
\label{eq:LOSS}
\end{equation}

Through the above auxiliary task, i.e. estimating the disparity of EPI corresponding to scene point after refocusing, the feature representation of EPI can be learned, so as to achieve self-supervised network training. Finally, the obtained self-supervised model is fine-tuned on the original depth estimation dataset to get the final depth estimation model. In addition, refocusing also extends the range of epipolar slope, which makes the network adapt to complex scenes and perform better in real scenarios, especially in occlusions.

\section{EPI-Stack Based CNN Architecture}
EPIs can achieve high estimation accuracy because of their special linear structure. However, traditional EPIs are limited by a single spatial dimension and are susceptible to noise when estimating scene depth. In this section, we focus on how to improve the original EPI input mode, build EPI-Stack, and improve the anti-noise performance. Then, based on the current input, we propose the corresponding network to process EPI-Stack.

\subsection{EPI-Stack Pairs}
The EPI is the 2D slice from 4D light field. We use $(u, v, x, y)$ to represent the 4D light field coordinates shown in Figure~\ref{fig:EPI-Stack}, where $(u, v)$ represents the directional dimension and $(x, y)$ represents the spatial dimension. Here, by fixing $v$ and $y$, or $u$ and $x$, respectively, the EPI in the horizontal or vertical direction can be obtained. Due to the limitation of a single space dimension, it is easily affected by the local spatial position. Therefore, extracting depth from EPI is easily disturbed by noise in the scene, and it is difficult to extract effective linear structure features~\cite{Intrinsics}. Existing methods generally use different directions of EPI as input to improve the robustness of depth estimation. However, due to the limitation of spatial dimension, EPI in different directions only has a single common spatial pixel point, which not only reduces the estimation efficiency (EPI in different directions can only estimate the depth of the central pixel point), but also makes it difficult for different directions of input to effectively complement each other, resulting in redundant information. Moreover, the estimation accuracy of noise scenes cannot be improved effectively. To solve these problems, we propose to increase the common spatial location points of EPI in different directions. Here we use the classical horizontal and vertical input modes. No matter horizontal or vertical EPI, there is only a sample of a single spatial location in its perspective dimension, so EPI in different directions has only a single common intersection. Therefore, our method is to increase the spatial resolution of horizontal and vertical EPI by increasing the sampling of spatial pixel points in the perspective dimension, so that the common area of EPI in both directions is maximized. As shown in Figure~\ref{fig:EPI-Stack}, in order to increase the spatial location points common to EPIs in different directions, EPIs with different spatial sampling points are stacked in the directional dimension, that is, from $(u, x)$ to $(u \cdot y, x)$, and from $(v, y)$ to $(v \cdot x, y)$. This representation is called EPI-Stack in this paper. In this way, we improve the spatial resolution of the polar plane image, and it provides more common space locations for the horizontal and vertical EPI in the perspective dimension, which increases the spatial constraint of EPI and is robust to noise.

\begin{figure}[t]
\begin{center}
\includegraphics[width=0.9\linewidth, height=0.6\linewidth, scale=1]{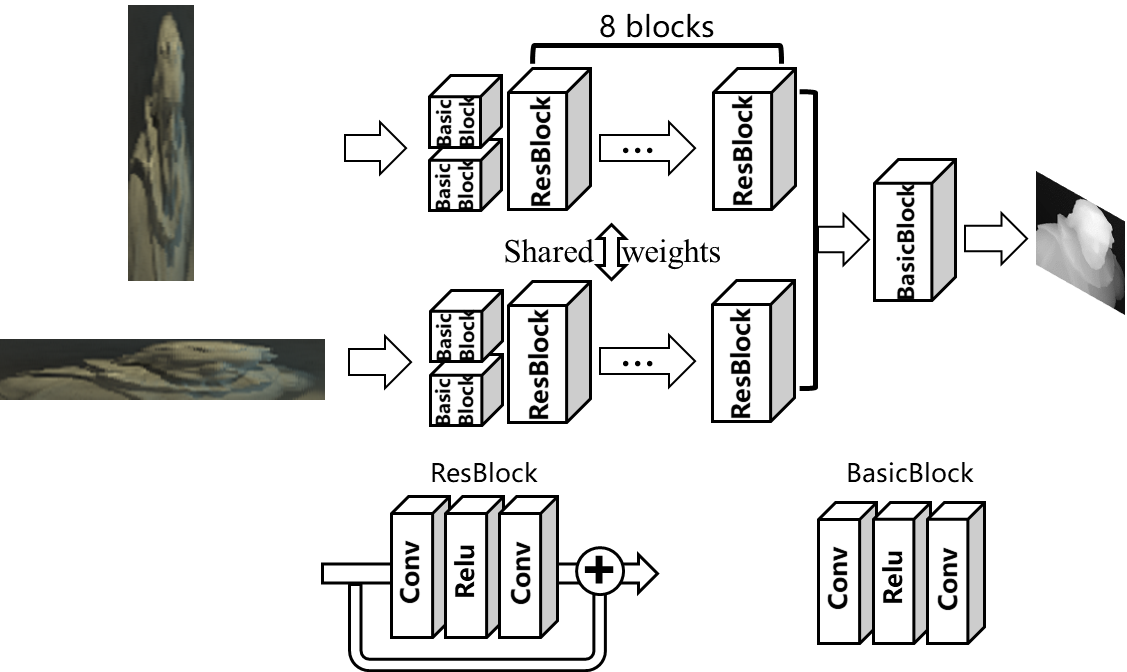}
\end{center}
   \caption{An overview of the proposed network architecture based on EPI-Stack.}
\label{fig:network}
\end{figure}

\subsection{Our CNN Architecture}
For this stacked EPI image, the corresponding network structure is designed for feature extraction. As shown in Figure~\ref{fig:network}, the stacked EPI images in both horizontal and vertical directions are input into the Siamese network, where 'Conv-ReLU-Conv' is used as the basic module for feature extraction. Since the spatial location sampling points of EPI-Stack are the same in both horizontal and vertical directions, the weight of the two-branch network is shared here. Compared with traditional EPI, the size of EPI-Stack proposed in this paper is too large. In order to quickly extract the spatial structure of EPI and learn the subtle changes of polar slope, we add feature extraction modules with different scales at the front of the network, using convolution layers with dimensions of $5 \times 5$ and $3 \times 3$ respectively, and set the convolution stride of stacked perspective dimension to 3 and the spatial dimension to 1. Several residual modules are used to extract the deep structure features of the scene, and then the disparity features of the scene are obtained by concatenating the outputs of the two-branch network. Here, except for the first basic module, all are convolution filters with size of $2 \times 2$ and stride 1. 

\section{Experiments}
\subsection{Implementation Details}
According to the self-supervised learning method in Section~\ref{sec3}, we use the Siamese network based on EPI-Stack to estimate the disparity values of the scene before and after focusing and compute the disparity shifts. Using the loss function in Eq.~\ref{eq:LOSS}, we can achieve the self-supervised training of the network. Then we use mean absolute error as the loss function to fine tune the Siamese network, as shown in Eq.~\ref{eq:L}.

\begin{equation}
L({y_i},{\hat y_i}) = \frac{1}{M}\sum\limits_{i = 1}^M {\left| {{y_i} - {{\hat y}_i}} \right|} 
\label{eq:L}
\end{equation}

Given $M$ EPI-Stack pairs in mini-batch, we denote the ground truth disparity of the i-th image as $y_{i}$, and the predicted value from the network is ${{\hat y}_i}$. The above networks are trained on the 4D HCI dataset ~\cite{dataset2016}, which provides 24 well-designed scenarios and corresponding disparity maps for quantitative evaluation. Each scene has $512 \times 512$ spatial resolutions and $9 \times 9$ angular resolutions. In this experiment, 16 scenarios were used for training and the rest for testing. Considering the limited number of training samples, we randomly extracted EPI-Stack samples of size $(9 \cdot 25) \times 25 \times 3$ from Horizontal and vertical views for training. We still use $(9 \cdot 512) \times 512 \times 3$ as the input size for testing. In this way, we not only get enough samples for training, but also use the whole scene for testing, which significantly improves the estimation efficiency compared to the existing EPI-based methods. In addition, we use the Keras framework to implement the network with a Nvidia 1080 Ti GPU. We set the batch size to 80, use the RMSprop optimizer, and set the weight decay rate to $10^{-5}$. Note that the proposed network is end-to-end trained and does not use subsequent complex processing measures.

\begin{figure}[t]
\begin{center}
\includegraphics[width=1.05\linewidth, height=0.7\linewidth, trim=12 0 0 0, clip]{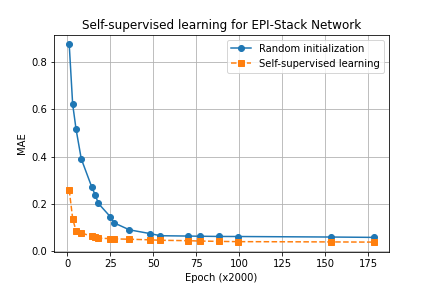}
\end{center}
\caption{Validation curve of self-supervised learning.}
\label{fig:curve}
\end{figure}

We use the bad pixel ratio (BadPix), which denotes the percentage of pixels with 0.07 error value, the Mean Square Error (MSE), and Peak Signal to Noise Ratio (PSNR(dB)) to evaluate the performance of our approach. For the predicted disparity map $d$, the ground truth $gt$, and the scene region $N$, its BadPix is defined as,

\begin{equation}
{\rm{BadPix = }}\frac{{\left| {\left\{ {x \in N:\left| {d(x) - gt(x)} \right| > 0.07} \right\}} \right|}}{{\left| N \right|}}
\label{eq:BP}
\end{equation}
\\
and MSE is defined as,
\begin{equation}
{\rm{MSE}} = \frac{{\sum\limits_{x \in N} {{{\left( {d(x) - gt(x)} \right)}^2}} }}{{\left| N \right|}} \times 100
\label{eq:MSE}
\end{equation}
\\
Lower scores are better for BadPix and MSE.

To evaluate the effectiveness of our approach in real-world scenarios, this paper conducts a qualitative experimental analysis using the Stanford Light Field Archive real-world dataset~\cite{Stanford2016}. These data are captured with the Lytro Illum handheld camera and contain a variety of complex real scenes, which can be used to better evaluate the performance of depth estimation.

\begin{figure}[t]
\begin{center}
\subfigure[]{
\begin{minipage}[t]{1\linewidth}
\centering
\includegraphics[width=1\linewidth, height=0.3\linewidth]{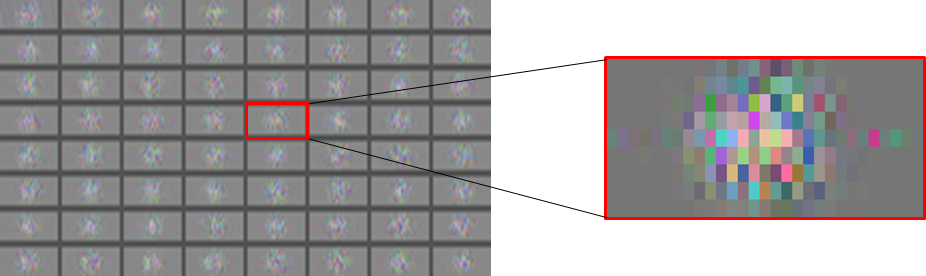}
\end{minipage}
}

\subfigure[]{
\begin{minipage}[t]{1\linewidth}
\centering
\includegraphics[width=1\linewidth, height=0.3\linewidth]{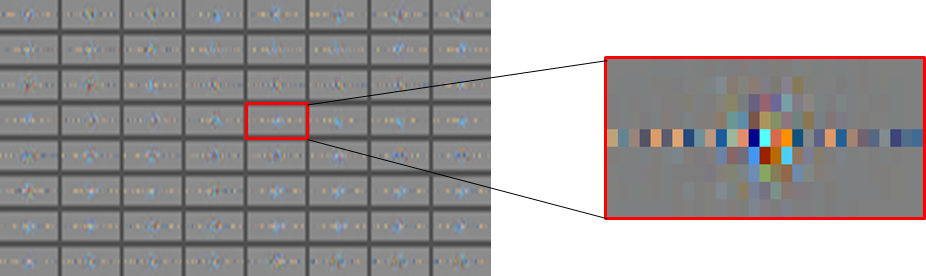}
\end{minipage}
}
\end{center}
\caption{Comparison of network visualization filters. (a) Random initialization. (b) Self-supervised learning.}
\label{fig:visual}
\end{figure}

\begin{figure}[t]
\begin{center}
\includegraphics[width=1\linewidth, height=0.7\linewidth, trim=0 15 0 0,clip]{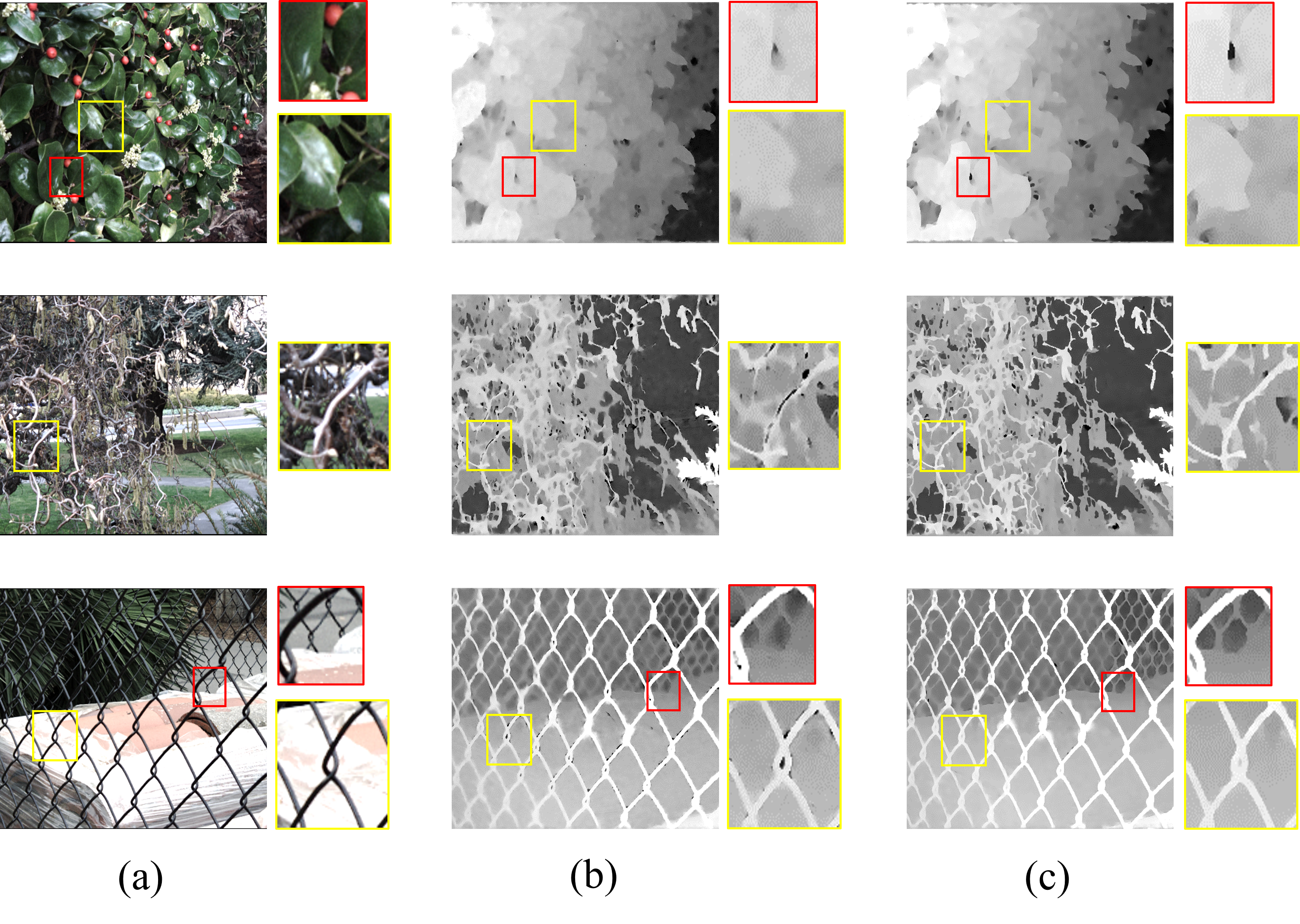}
\end{center}
\caption{Comparison of self-supervised depth estimation results. (a) Scene. (b) Random initialization. (c) Self-supervised learning.}
\label{fig:self-depth}
\end{figure}

\subsection{Ablation Analysis}
{\bf The effect of self-supervised learning on depth estimation.} We illustrate the validation of the proposed self-supervised depth estimation by comparing the loss curves. From the validation of the loss curve in Fig~\ref{fig:curve}, it can be seen that the self-supervised depth estimation method can accelerate the convergence of the original network compared with the randomly initialized network model. At the same time, we visualize the network convolution kernel before and after self-supervised learning, which is illustrated with a single EPI. Compared with the randomly initialized model (Fig~\ref{fig:visual}(a), the self-supervised depth estimation method can compute the disparity shift according to the slight change of EPI slope before and after refocusing, which enables the network filter to learn a more efficient linear representation of the EPI (Fig~\ref{fig:visual}(b)). These visual filters are extracted from the last layer of a single network branch.

\begin{figure*}[t]
\setlength{\abovecaptionskip}{-0.1cm}
\setlength{\belowcaptionskip}{0.5cm}
\begin{center}

\includegraphics[width=0.95\linewidth, height=0.6\linewidth, trim=0 8 0 0,clip]{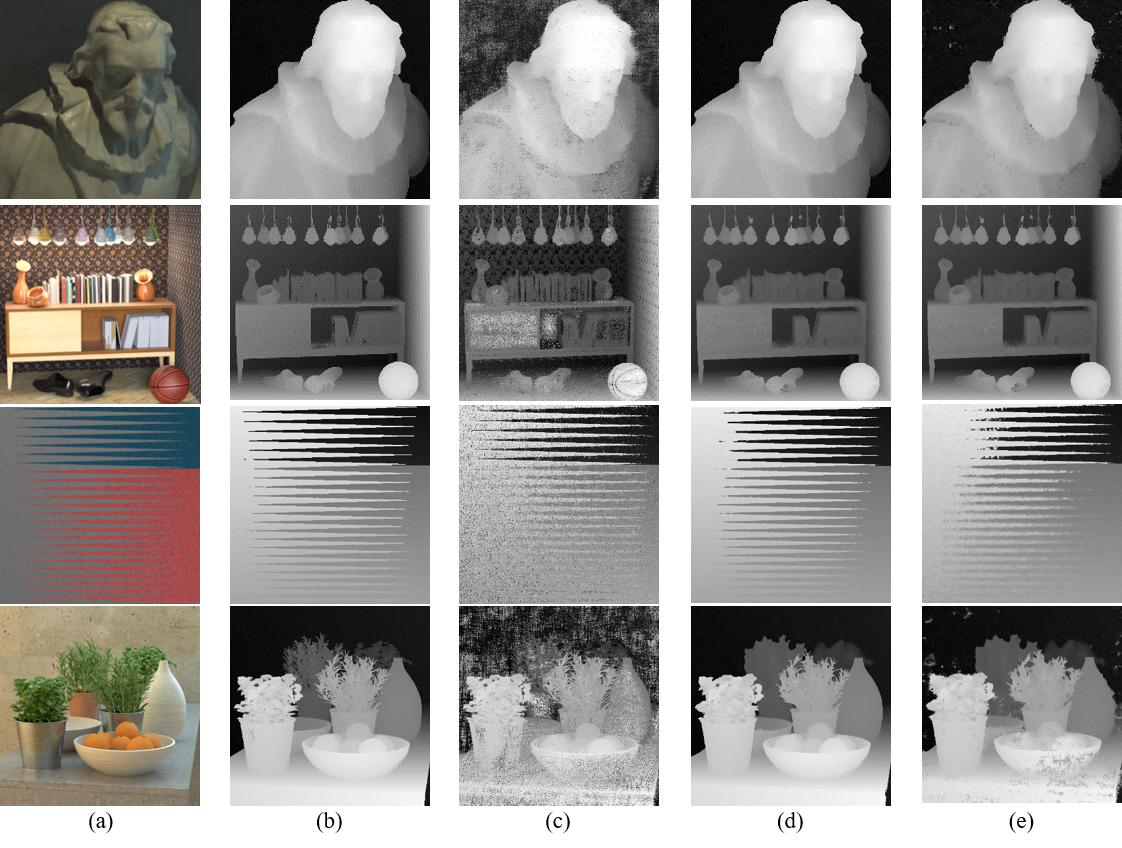}
\end{center}
\caption{Comparison of disparity maps on noisy scenes. (a) Scene. (b) EPI\_ORM~\cite{EPI-ORM}. (c) EPI\_ORM (noise). (d) EPI-Stack. (e) EPI-Stack (noise).}
\label{fig:noise}
\end{figure*}

We use the self-supervised method to estimate the disparity map in Stanford real scene. In the experiment, we select some complex occlusion scenes for qualitative comparison. As shown in Figure~\ref{fig:self-depth}, it can be found from the first row that compared with random initialization, the self-supervised method can effectively estimate the depth discontinuities and obtain leaves with clear boundaries. Similarly, in the complex scene of the second row, our method is robust to the interference of complex background and recovers the branches in the foreground. The last row is a challenging scene with multiple occlusion. The wire meshes in the fore- and 
back-ground overlap each other and contain a lot of details at the intersection of the meshes. We can find that before using self-monitoring training, our network can recover the scene contour, but in the detail region, such as the region marked by yellow line, the fore- and back-ground interfere with each other. So it is difficult to extract effective depth features. After training with self-supervised learning, the details of the scene are restored. In addition, in the region marked by red line, our approach not only restores foreground information, but also enables more accurate estimates of backgrounds at different depths.

\begin{figure*}[!t]
\begin{center}
\includegraphics[width=1\linewidth, height=0.7\linewidth, trim= 0 9 0 0, clip]{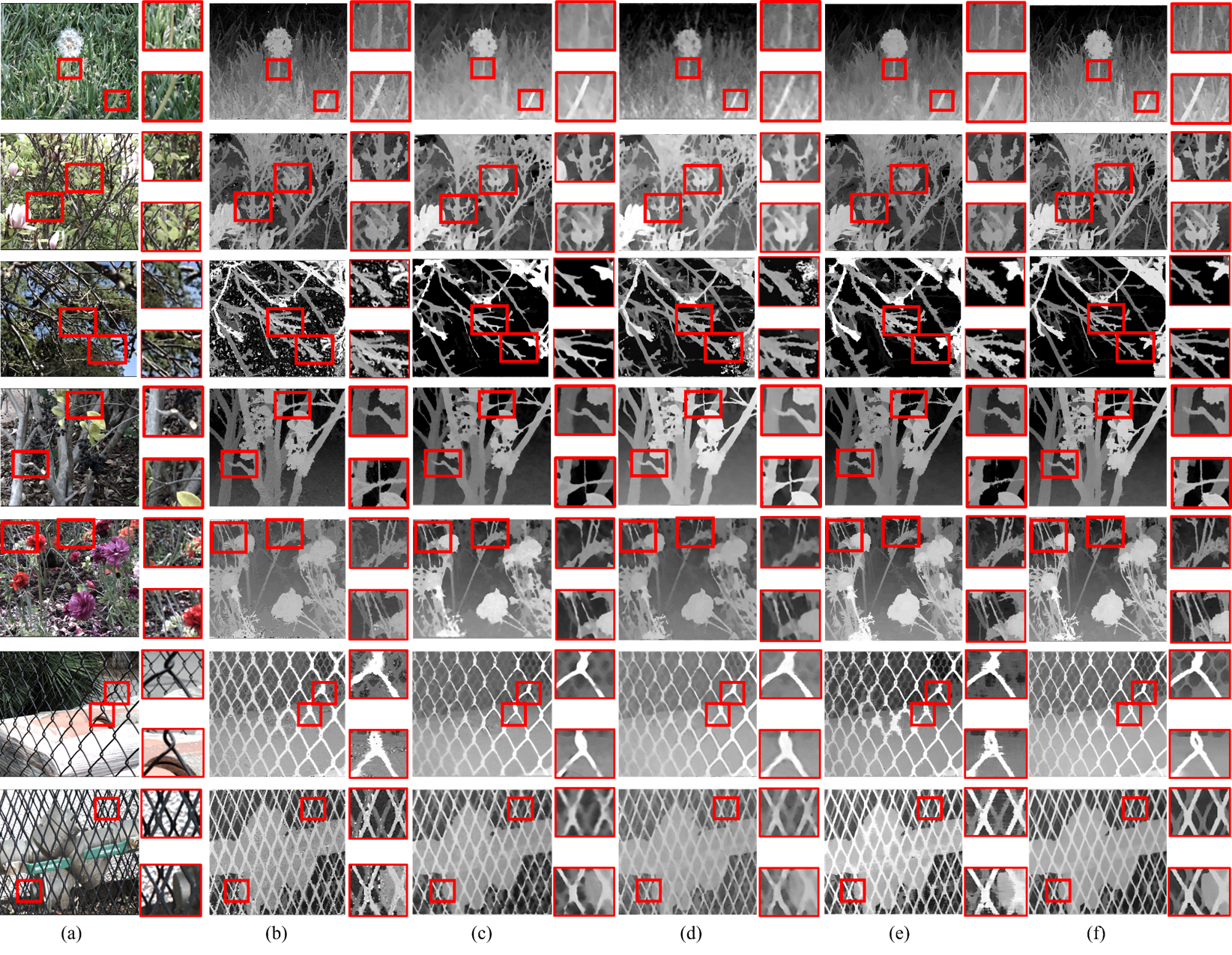}
\end{center}
\caption{Qualitative results on Stanford Light Field Data. (a) Scene. (b) CAE~\cite{Williem2016}. (c) EPINET~\cite{EPINET}. (d) EPI-Shift~\cite{EPI-shift}. (e) EPI\_ORM~\cite{EPI-ORM}. (f) Ours.}
\label{fig:Stanford}
\end{figure*}

\begin{table*}
\caption{Quantitative results for current state-of-the-art methods averaged on synthetic data. The best results are in bold and underlined.}
\small 
\begin{center}
\renewcommand\arraystretch{1.2}
\setlength{\tabcolsep}{4mm}{
\begin{tabular}{|l|c c c c c c c|}

\hline
Metric & CAE~\cite{Williem2016} & SPO~\cite{SPO} & EPN~\cite{EPI-patch} & EPINET~\cite{EPINET} & EPI-Shift~\cite{EPI-shift} & EPI\_ORM~\cite{EPI-ORM} & Ours \\
\hline\hline
BadPix & 9.016 & 7.475 & 7.060 & 6.057 & 11.471 & {\bf 5.660} & \underline{5.793} \\
MSE & 2.797 & 2.939 & 3.257 & 2.105 & 2.979 & \underline{1.393} & {\bf 1.320} \\
PSNR(dB) & 29.387 & 29.503 & 26.462 & 31.892 & 29.345 & \underline{31.906} & {\bf 32.209} \\
Time & 13min 52s & 34min 5s & 2min 58s & \underline{2.6s} & 22.6s & 54.2s & {\bf 2.1s}\\
GPU  & \XSolidBrush & \XSolidBrush & \CheckmarkBold & \CheckmarkBold & \CheckmarkBold & \CheckmarkBold & \CheckmarkBold \\
\hline

\end{tabular} }
\end{center}

\label{table:tab}
\end{table*}

{\bf The effect of EPI-Stack input mode.} To verify the advantage of EPI-Stack input mode in noisy scenes, we add extra Gaussian noise to the original scene and compare it with the disparity before adding noise. Here we compare the recent method EPI\_ORM~\cite{EPI-ORM} based on traditional EPI input mode, which achieves the state-of-the-art in multiple scenarios through the relation network. In this experiment, to compare the effect of different input modes on performance, we implement end-to-end training on the original network without self-supervised training. By comparing the estimation results of two input modes in Fig~\ref{fig:noise}, we find that the traditional EPI input mode can achieve high estimation accuracy in the original scene. However the disparity map estimated after adding noise is accompanied by obvious blurring, which is also the shortage of the current traditional EPI input mode. By contrast, our proposed approach with EPI-Stack still yields clear results. This is because the EPI-Stack input method increases the sampling of spatial pixel points, which enables EPI in different directions to have larger public areas and introduces stronger spatial constraints. Therefore, this input mode can improve the robustness of depth estimation for noise scenes.

\subsection{Comparison with the state-of-the-art}
We compare our approach with state-of-the-art methods: CAE~\cite{Williem2016}, SPO~\cite{SPO}, EPN~\cite{EPI-patch}, EPINET~\cite{EPINET}, EPI-Shift~\cite{EPI-shift}, and EPI\_ORM~\cite{EPI-ORM}. We use the training scene (boxes, cotton, dino, and sideboard) from 4D HCI Benchmark for quantitative comparison. Table~\ref{table:tab} shows quantitative results for these state-of-the-art methods averaged on synthetic scenes. It can be seen that our method achieves second best result with BadPix metric and the best performance with MSE and PSNR metrics. This is because our method can not only get effective refocusing cues, but also this special EPI-Stack input mode increases the number of common spatial sampling points, improves the robustness of depth estimation, and makes the disparity map smoother. In addition, we provide a comparison of average runtimes and an indication of whether the GPU implementation is used. Compared with other methods, our approach achieves higher estimation accuracy and efficiency.

Figure~\ref{fig:Stanford} provides a qualitative comparison of the current method on Stanford light field data. Note that we extract the light field data of $9 \times 9$ viewing angle from the real data to test. We set the parameter label of the traditional method CAE~\cite{Williem2016} to 75. The real scene parallax range is [-1, 1] by default. Qualitative comparison shows that the traditional CAE method is susceptible to noise interference in the real scene, which reduces the estimation accuracy. This method also takes a long time to compute. Methods EPINET~\cite{EPINET} and EPI-Shift~\cite{EPI-shift} both use sub-aperture images as input. Although the scene contour is recovered, the local details are not well estimated and are obviously blurred. EPI\_ORM~\cite{EPI-ORM} method uses the traditional EPI input mode and restores the details of the scene better. However, due to the limitations of synthetic scene training, this method does not adapt to complex disparity changes in real scenes. Therefore the occlusion boundaries of scenes cannot be restored in Figure~\ref{fig:Stanford}. Our method is not trained in a real scene and still obtains high disparity map. Through self-supervised learning based on refocusing, our approach extends the disparity range of EPI, extracts effective refocusing cues, and achieves higher estimation accuracy. Especially in discontinuous and multi-occluded regions, our method obtains clear contours and preserves the foreground and background boundaries.

\section{Conclusion}
In this paper, we propose a self-supervised learning framework for light field depth estimation by analyzing EPI disparity shift. In this way, we combine the light field refocusing with CNN to get better estimation performance, especially in occlusions. In addition, we also propose the EPI-Stack input mode, which increases the spatial sampling points of EPI in different directions. This input mode can further improve the estimation efficiency and performance of EPI in noise scenes. Experimental results on synthetic and real scenes show that our 
approach obtains more accurate disparity map than those produced by competing the state-of-the-arts. And contours and occluded boundaries of real scenes are restored better.


Acknowledgments.This work was supported by the National Natural Science Foundation of China, No. 61876057.



\bibliographystyle{IEEEtran}
\bibliography{egbib}

\begin{thebibliography}{10}
\providecommand{\url}[1]{#1}
\csname url@samestyle\endcsname
\providecommand{\newblock}{\relax}
\providecommand{\bibinfo}[2]{#2}
\providecommand{\BIBentrySTDinterwordspacing}{\spaceskip=0pt\relax}
\providecommand{\BIBentryALTinterwordstretchfactor}{4}
\providecommand{\BIBentryALTinterwordspacing}{\spaceskip=\fontdimen2\font plus
\BIBentryALTinterwordstretchfactor\fontdimen3\font minus
  \fontdimen4\font\relax}
\providecommand{\BIBforeignlanguage}[2]{{%
\expandafter\ifx\csname l@#1\endcsname\relax
\typeout{** WARNING: IEEEtran.bst: No hyphenation pattern has been}%
\typeout{** loaded for the language `#1'. Using the pattern for}%
\typeout{** the default language instead.}%
\else
\language=\csname l@#1\endcsname
\fi
#2}}
\providecommand{\BIBdecl}{\relax}
\BIBdecl

\bibitem{Ng}
R.~Ng, M.~Levoy, G.~Duval, M.~Horowitz, and P.~Hanrahan, ``Light field
  photography with a hand-held plenoptic camera,'' \emph{Computer Science
  Technical Report}, vol.~2, no.~11, pp. 1--11, 2005.

\bibitem{Levoy}
M.~Levoy and P.~Hanrahan, ``Light field rendering,'' \emph{In Proceedings of
  the 23rd Annual Conference on Computer Graphics and Interactive Techniques},
  pp. 31--42, 1996.

\bibitem{Feng2018}
M.~Feng, Y.~Wang, J.~Liu, L.~Zhang, H.~F.~M. Zaki, and A.~Mian, ``Benchmark
  dataset and method for depth estimation from light field images,'' \emph{IEEE
  Transactions on Image Processing}, vol.~27, no.~7, pp. 3586--3598, 2018.

\bibitem{EPI-volume}
S.~Heber, W.~Yu, and T.~Pock, ``Neural epi-volume networks for shape from light
  field,'' \emph{In Proceedings of International Conference on Computer Vision
  (ICCV)}, pp. 2271--2279, 2017.

\bibitem{EPI-shift}
T.~Leistner, H.~Schilling, R.~Mackowiak, S.~Gumhold, and C.~Rother, ``Learning
  to think outside the box: Wide-baseline light field depth estimation with
  epi-shift,'' \emph{In Proceedings of 2019 International Conference on 3D
  Vision (3DV)}, pp. 249--257, 2019.

\bibitem{EPI-ORM}
K.~Li, J.~Zhang, R.~Sun, X.~Zhang, and J.~Gao, ``Epi-based oriented relation
  networks for light field depth estimation,'' \emph{In Proceedings of British
  Machine Vision Conference (BMVC)}, 2020.

\bibitem{dataset2016}
K.~Honauer, O.~Johannsen, D.~Kondermann, and B.~Goldluecke, ``A dataset and
  evaluation methodology for depth estimation on 4d light fields,'' \emph{In
  Proceedings of Asian Conference on Computer Vision (ACCV)}, pp. 19--34, 2016.

\bibitem{dataset2013}
S.~Wanner, S.~Meister, and B.~Goldluecke, ``Datasets and benchmarks for densely
  sampled 4d light fields,'' \emph{In Proceedings of Vision, Modelling \&
  Visualization (VMV)}, pp. 225--226, 2013.

\bibitem{Tao2013}
M.~W. Tao, S.~Hadap, J.~Malik, and R.~Ramamoorthi, ``Depth from combining
  defocus and correspondence using light-field cameras,'' \emph{In Proceedings
  of International Conference on Computer Vision (ICCV)}, pp. 673--680, 2013.

\bibitem{Tao2017}
M.~W. Tao, P.~P. Srinivasan, S.~Hadap, S.~Rusinkiewicz, J.~Malik, and
  R.~Ramamoorthi, ``Shape estimation from shading, defocus, and correspondence
  using light-field angular coherence,'' \emph{IEEE Transactions on Pattern
  Analysis and Machine Intelligence}, vol.~39, no.~3, pp. 546--560, 2017.

\bibitem{Wanner2014}
S.~Wanner and B.~Goldluecke, ``Variational light field analysis for disparity
  estimation and super-resolution,'' \emph{IEEE Transactions on Pattern
  Analysis and Machine Intelligence}, vol.~36, no.~3, pp. 606--619, 2014.

\bibitem{Williem2016}
W.~Williem and I.~K. Park, ``Robust light field depth estimation for noisy
  scene with occlusion,'' \emph{In Proceedings of IEEE Conference on Computer
  Vision and Pattern Recognition (CVPR)}, pp. 4396--4404, 2016.

\bibitem{focal-stack}
H.~Lin, C.~Chen, S.~B. Kang, and J.~Yu, ``Depth recovery from light field using
  focal stack symmetry,'' \emph{In Proceedings of International Conference on
  Computer Vision (ICCV)}, pp. 3451--3459, 2015.

\bibitem{RankIQA}
X.~Liu, J.~V.~D. Weijer, and A.~D. Bagdanov, ``Rankiqa: Learning from rankings
  for no-reference image quality assessment,'' \emph{In Proceedings of
  International Conference on Computer Vision (ICCV)}, pp. 1040--1049, 2017.

\bibitem{RankIQA-2019}
------, ``Exploiting unlabeled data in cnns by self-supervised learning to
  rank,'' \emph{IEEE Transactions on Pattern Analysis and Machine
  Intelligence}, vol.~41, no.~8, pp. 1862--1878, 2019.

\bibitem{EPI-patch}
Y.~Luo, W.~Zhou, J.~Fang, L.~Liang, H.~Zhang, and G.~Dai, ``Epi-patch based
  convo-lutional neural network for depth estimation on 4d light field,''
  \emph{In Proceedings of In-ternational Conference on Neural Information
  Processing (ICONIP)}, pp. 642--652, 2017.

\bibitem{EPINET}
C.~Shin, H.~Jeon, Y.~Yoon, I.~S. Kweon, and S.~J. Kim, ``Epinet: A
  fully-convolutional neural network using epipolar geometry for depth from
  light field images,'' \emph{In Proceedings of IEEE Conference on Computer
  Vision and Pattern Recognition (CVPR)}, pp. 4748--4757, 2018.

\bibitem{Jeon2019}
H.~G. Jeon, J.~Park, G.~Choe, J.~Park, Y.~Bok, Y.~W. Tai, and I.~S. Kweon,
  ``Depth from a light field image with learning-based matching costs,''
  \emph{IEEE Transactions on Pattern Analysis and Machine Intelligence},
  vol.~41, no.~2, pp. 297--310, 2019.

\bibitem{Wang2015}
T.~C. Wang, A.~Efros, and R.~Ramamoorthi, ``Occlusion-aware depth estimation
  using light-field cameras,'' \emph{In Proceedings of International Conference
  on Computer Vision (ICCV)}, pp. 3487--3495, 2015.

\bibitem{saliency}
M.~Zhang, W.~Ji, Y.~Piao, J.~Li, Y.~Zhang, S.~Xu, and H.~Lu, ``Lfnet: Light
  field fusion network for salient object detection,'' \emph{IEEE Transactions
  on Image Processing}, vol.~29, pp. 6276--6287, 2020.

\bibitem{Super-Resolution}
S.~Zhang, Y.~Lin, and H.~Sheng, ``Residual networks for light field image
  super-resolution,'' \emph{In Proceedings of IEEE Conference on Computer
  Vision and Pattern Recognition (CVPR)}, pp. 11\,038--11\,047, 2019.

\bibitem{Jeon2015}
H.~G. Jeon, J.~Park, G.~Choe, J.~Park, Y.~Bok, Y.~W. Tai, and I.~S. Kweon,
  ``Accurate depth map estimation from a lenslet light field camera,'' \emph{In
  Proceedings of IEEE Conference on Computer Vision and Pattern Recognition
  (CVPR)}, pp. 1547--1555, 2015.

\bibitem{Unet2015}
O.~Ronneberger, P.~Fischer, and T.~Brox, ``U-net: Convolutional networks for
  biomedical image segmentation,'' \emph{In Proceedings of International
  Conference on Medical Image Computing and Computer-Assisted Intervention
  (MICCAI)}, pp. 234--241, 2015.

\bibitem{Wanner2013}
S.~Wanner, C.~Straehle, and B.~Goldluecke, ``Globally consistent multi-label
  assignment on the ray space of 4d light fields,'' \emph{In Proceedings of
  IEEE Conference on Computer Vision and Pattern Recognition (CVPR)}, pp.
  1011--1018, 2013.

\bibitem{SPO}
S.~Zhang, H.~Sheng, C.~Li, J.~Zhang, and Z.~Xiong, ``Robust depth estimation
  for light field via spinning parallelogram operator,'' \emph{Computer Vision
  and Image Understanding}, vol. 145, pp. 148--159, 2016.

\bibitem{EPI2017}
Y.~Zhang, H.~Lv, Y.~Liu, H.~Wang, X.~Wang, Q.~Huang, X.~Xiang, and Q.~Dai,
  ``Light-field depth estimation via epipolar plane image analysis and locally
  linear embedding,'' \emph{IEEE Transactions on Circuits and Systems for Video
  Technology}, vol.~27, no.~4, pp. 739--747, 2017.

\bibitem{Intrinsics}
A.~Alperovich, O.~Johannsen, M.~Strecke, and B.~Goldluecke, ``Light field
  intrinsics with a deep encoder-decoder network,'' \emph{In Proceedings of
  IEEE Conference on Computer Vision and Pattern Recognition (CVPR)}, pp.
  9145--9154, 2018.

\bibitem{Heber-shape}
S.~Heber and T.~Pock, ``Convolutional networks for shape from light field,''
  \emph{In Proceedings of IEEE Conference on Computer Vision and Pattern
  Recognition (CVPR)}, pp. 3746--3754, 2016.

\bibitem{Heber-Unet}
S.~Heber, Y.~Wei, and T.~Pock, ``U-shaped networks for shape from light
  field,'' \emph{In Proceedings of British Machine Vision Conference (BMVC)},
  2016.

\bibitem{Sun2016}
X.~Sun, Z.~Xu, N.~Meng, E.~Y. Lam, and H.~K.~H. So, ``Data-driven light field
  depth estimation using deep convolutional neural networks,'' \emph{In
  Proceedings of International Joint Conference on Neural Networks (IJCNN)},
  pp. 367--374, 2016.

\bibitem{EPI-Patch2018}
W.~Zhou, L.~Liang, H.~Zhang, A.~Lumsdaine, and L.~Lin, ``Scale and orientation
  aware epi-patch learning for light field depth estimation,'' \emph{In
  Proceedings of International Conference on Pattern Recognition (ICPR)}, pp.
  2362--2367, 2018.

\bibitem{Tao2015}
M.~W. Tao, P.~P. Srinivasan, J.~Malik, S.~Rusinkiewicz, and R.~Ramamoorthi,
  ``Depth from shading, defocus, and correspondence using light-field angular
  coherence,'' \emph{In Proceedings of IEEE Conference on Computer Vision and
  Pattern Recognition (CVPR)}, pp. 1940--1948, 2015.

\bibitem{focal-cue}
W.~Zhou, E.~Zhou, Y.~Yan, L.~Lin, and A.~Lumsdaine, ``Learning depth cues from
  focal stack for light field depth estimation,'' \emph{In Proceedings of
  International Conference on Image Processing (ICIP)}, pp. 1074--1078, 2019.

\bibitem{Doersch}
C.~Doersch, A.~Gupta, and A.~A. Efros, ``Unsupervised visual representation
  learning by context prediction,'' \emph{In Proceedings of IEEE International
  Conference on Computer Vision (ICCV)}, pp. 1422--1430, 2015.

\bibitem{Pathak}
D.~Pathak, P.~Krähenbühl, J.~Donahue, T.~Darrell, and A.~A. Efros, ``Context
  encoders: Feature learning by inpainting,'' \emph{In Proceedings of IEEE
  Conference on Computer Vision and Pattern Recognition (CVPR)}, pp.
  2536--2544, 2016.

\bibitem{Larsson}
G.~Larsson, M.~Maire, and G.~Shakhnarovich, ``Colorization as a proxy task for
  visual understanding,'' \emph{In Proceedings of IEEE Conference on Computer
  Vision and Pattern Recognition (CVPR)}, pp. 840--849, 2017.

\bibitem{Unsupervised-Depth2016}
R.~Garg, V.~K.~B. G, G.~Carneiro, and I.~Reid, ``Unsupervised cnn for single
  view depth estimation: Geometry to the rescue,'' \emph{In Proceedings of
  European Conference on Computer Vision (ECCV)}, 2016.

\bibitem{Unsupervised-Depth}
C.~Godard, O.~Aodha, and G.~J. Brostow, ``Unsupervised monocular depth
  estimation with left-right consistency,'' \emph{In Proceedings of IEEE
  Conference on Computer Vision and Pattern Recognition (CVPR)}, pp.
  6602--6611, 2017.

\bibitem{mono-depth-dig}
C.~Godard, O.~M. Aodha, M.~Firman, and G.~Brostow, ``Digging into
  self-supervised monocular depth estimation,'' \emph{In Proceedings of IEEE
  International Conference on Computer Vision (ICCV)}, pp. 3827--3837, 2019.

\bibitem{mono-depth}
J.~Watson, M.~Firman, G.~Brostow, and D.~Turmukhambetov, ``Self-supervised
  monocular depth hints,'' \emph{In Proceedings of IEEE International
  Conference on Computer Vision (ICCV)}, pp. 2162--2171, 2019.

\bibitem{Stanford2016}
A.~S. Raj, M.~Lowney, and R.~Shah, ``Light-field database creation and depth
  estimation,'' \emph{USA: Stanford University}, 2016.

\end{thebibliography}

\end{document}